\documentclass[10pt,twocolumn,letterpaper]{article}

\usepackage{iccv}
\usepackage{times}
\usepackage{epsfig}
\usepackage{graphicx}
\usepackage{amsmath}
\usepackage{amssymb}
\usepackage{comment}
\usepackage{soul}
\usepackage{pdfpages}

\usepackage{booktabs}
\usepackage{caption}
\usepackage{xcolor}
\usepackage{nicefrac}
\usepackage{subfig}
\usepackage{colortbl}

\usepackage{algorithm}
\usepackage{listings}
\usepackage{bm}
\usepackage{dsfont}
\usepackage[accsupp]{axessibility}

\usepackage{etoolbox}
\makeatletter
\AfterEndEnvironment{algorithm}{\let\@algcomment\relax}
\AtEndEnvironment{algorithm}{\kern2pt\hrule\relax\vskip3pt\@algcomment}
\let\@algcomment\relax
\newcommand\algcomment[1]{\def\@algcomment{\footnotesize#1}}
\renewcommand\fs@ruled{\def\@fs@cfont{\bfseries}\let\@fs@capt\floatc@ruled
  \def\@fs@pre{\hrule height.8pt depth0pt \kern2pt}%
  \def\@fs@post{}%
  \def\@fs@mid{\kern2pt\hrule\kern2pt}%
  \let\@fs@iftopcapt\iftrue}
\makeatother

\usepackage[pagebackref=true,breaklinks=true,colorlinks,bookmarks=false]{hyperref}

\usepackage[capitalize]{cleveref}
\crefname{section}{Sec.}{Secs.}
\Crefname{section}{Section}{Sections}
\Crefname{table}{Table}{Tables}
\crefname{table}{Tab.}{Tabs.}

\definecolor{baselinecolor}{gray}{.9}
\newcommand{\baseline}[1]{\cellcolor{baselinecolor}{#1}}

\newcommand{\uline}[1]{\ul{#1}}

\renewcommand*{\ie}{i.e.\@\xspace}
\renewcommand*{\eg}{e.g.\@\xspace}

\iccvfinalcopy %

\ificcvfinal\fi

\begin{document}

\title{Incremental Generalized Category Discovery}

\author{Bingchen Zhao
\quad  Oisin Mac Aodha \\ [0.3em]
University of Edinburgh \\
{\tt\small \url{https://bzhao.me/iNatIGCD}
}
}

\twocolumn[{%
\renewcommand\twocolumn[1][]{#1}%
\maketitle
\begin{center}
\centering
\captionsetup{type=figure}
    \vspace{-10pt}
    \includegraphics[width=0.70\linewidth]{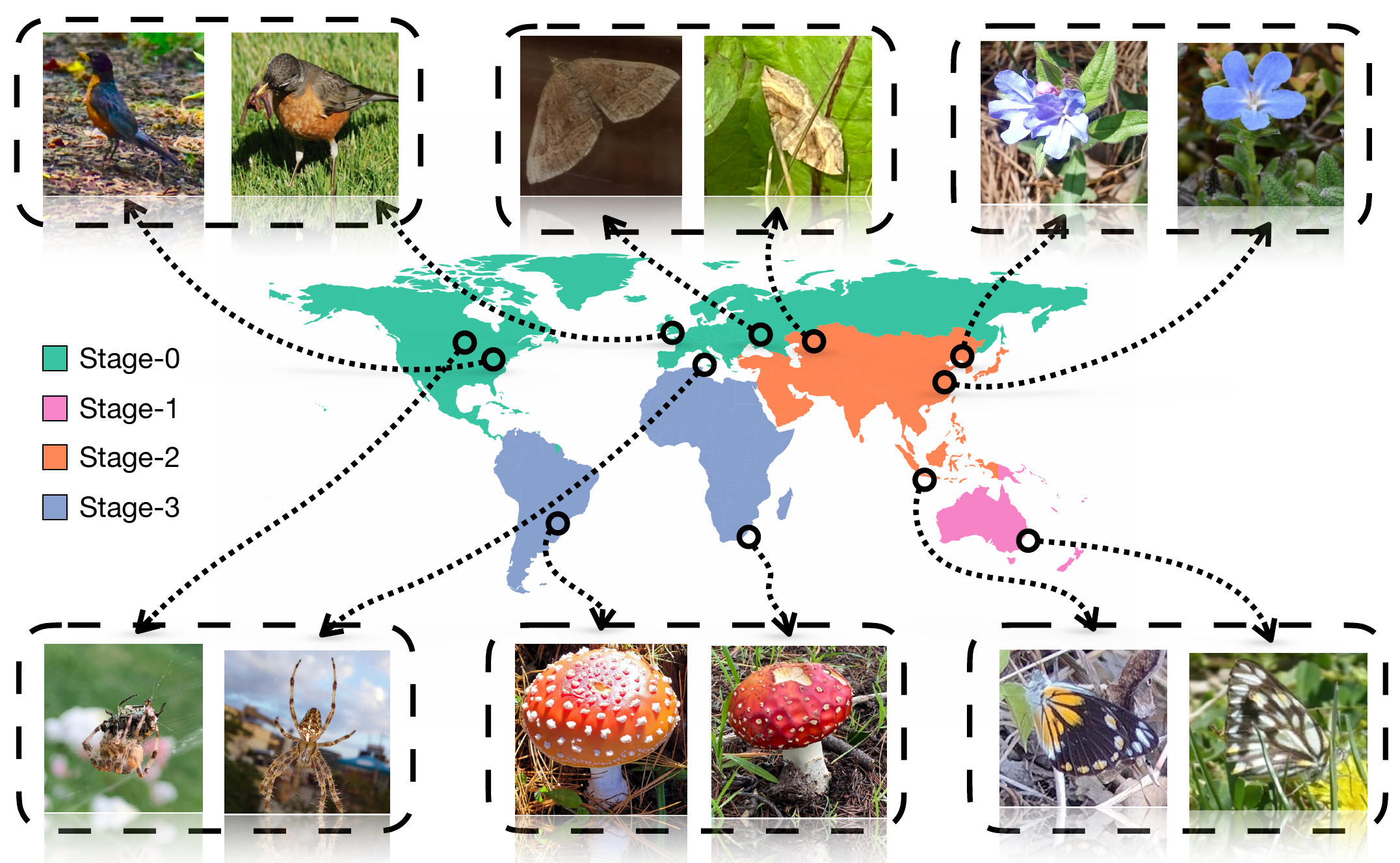}
\vspace{-5pt}
\caption{
We explore the task of discovering novel categories over time. 
Here we illustrate visual differences between the same categories across different continents, where the continent colors represent four different incremental learning stages from our new iNatIGCD dataset. 
Each box contains two instances of the same category, where the images are from different geographical locations.
We can see that the background and appearance can change dramatically across different locations.
}
\label{fig:species}
\end{center}
}]
\ificcvfinal\thispagestyle{empty}\fi

\begin{abstract} 
\vspace{-7pt}
We explore the problem of Incremental Generalized Category Discovery (IGCD). 
This is a challenging category-incremental learning setting where the goal is to develop models that can correctly categorize images from previously seen categories, in addition to discovering novel ones. 
Learning is performed over a series of time steps where the model obtains new labeled and unlabeled data, and discards old data, at each iteration. 
The difficulty of the problem is compounded in our generalized setting as the unlabeled data can contain images from categories that may or may not have been observed before. 
We present a new method for IGCD which combines non-parametric categorization with efficient image sampling to mitigate catastrophic forgetting.  
To quantify performance, we propose a new benchmark dataset named iNatIGCD that is motivated by a real-world fine-grained visual categorization task. 
In our experiments we outperform existing related methods. 
\end{abstract}
\vspace{-10pt}

\section{Introduction}
The wealth and complexity of visual information potentially observable by artificial systems deployed in the real world vastly exceeds the comparative simplicity of our carefully curated benchmark vision datasets. 
To operate safely and reliably in challenging environments, these systems need to be able to correctly recognize previously learned concepts, not confuse these known concepts with novel ones, and be able to differentiate novel concepts so that they can be grouped and efficiently learned. 
As humans, we excel at this type of flexible learning in such dynamic settings~\cite{ashby2005human,lake2015human}, and it is clear that we need to endow our artificial systems with similar desirable abilities.

In the context of visual categorization, there is a rich body of work that has moved beyond the traditional supervised setting into more open-ended learning paradigms. 
For example, in semi-supervised learning, in addition to labeled data, during training we also have access to unlabeled data depicting the same categories~\cite{zhu2005semi}. 
Other work has attempted to address more complex settings such as determining if an image contains a previously observed, or instead a novel category (\ie open-set recognition)~\cite{scheirer2012toward}, or learning to group images from novel categories (\ie novel category discovery)~\cite{han2019learning,troisemaine2023novel}. 
Recently, a more realistic and challenging setting termed Generalized Category Discovery (GCD) has been proposed~\cite{cao22,vaze2022generalized}. 
Here, unlabeled images can be from either previously seen \emph{or} from novel categories and the task is to develop a model that can both classify the previously seen categories and also discover novel ones.  

\begin{figure}[t]
\centering
    \includegraphics[width=1.0\columnwidth]{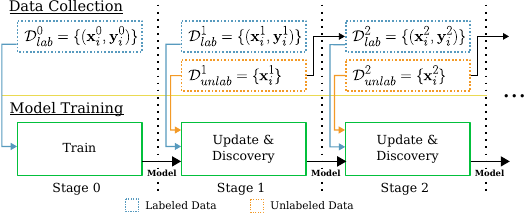}
\vspace{-20pt}
\caption{Overview of our incremental generalized category discovery setting. At each time step the model has access to labeled and unlabeled data and must simultaneously classify existing categories and discover new ones.  
As time progress, all previous data is no longer directly available. 
}
\label{fig:teaser}
\vspace{-10pt}
\end{figure}

In this work, we go one step further by exploring the generalized setting along the temporal, \ie category-incremental, dimension~\cite{thrun1995learning,rebuffi2017icarl}. 
We refer to this as {\bf I}ncremental {\bf G}eneralized {\bf C}ategory {\bf D}iscovery (IGCD). 
Here, learning progress over a series of stages (representing time steps), where at each stage we acquire a new set of unlabeled data that contains images from categories that may, or may not, have been previously observed (see \cref{fig:teaser}). 
At each stage, the goal when updating the model is to maintain performance on the previously observed categories, while also discovering novel ones. 
This must be achieved under the constraint that at each new stage the labeled data from earlier stages is no longer available for training. 

A small number of recent works have started to investigate category discovery in the context of incremental learning. 
For example, when only novel categories are present at each subsequent stage in the case of~\cite{roy2022class,joseph2022novel,liu2023large} or in the generalized setting in~\cite{zhang2022grow} where the unlabeled data at each each stage contains both old and novel categories.  
However, in contrast to~\cite{zhang2022grow}, in our IGCD setting we do not make the simplifying assumption that the unlabeled data at each stage contains \emph{all} the categories from the previous stages. 
More realistically, unlabeled data is not guaranteed to contain all previously seen categories which makes the task even harder for the incremental learner. 
We show that combining a state-of-the-art GCD method with incremental learning techniques can work, but results in forgetting as time progresses.  
We address these issues via a novel IGCD method that combines a non-parametric classifier with a density-based sampling mechanism that efficiently enables the selection of informative examples for both classification and past memorization.

Benchmarking in the incremental discovery setting to date has largely been restricted to artificial category-based splits of conventional image categorization datasets~\cite{cifar,WahCUB_200_2011,cars,fgvc_aircraft}. 
To encourage future progress on IGCD, we also present a new dataset called iNatIGCD. 
iNatIGCD more faithfully simulates a real-world fine-grained incremental learning setting. 
It is motivated by a real-world visual categorization use-case whereby the categories present at different stages are selected using spatio-temporal metadata from the community science platform iNaturalist~\cite{iNatWeb}. 
As a result, iNatIGCD naturally includes challenges such as appearance shifts, in addition to both fine and coarse-grained differences between old and novel categories present in images sourced from different geographical regions.

In summary, we present the following contributions:  
(i) A new approach for IGCD that combines non-parametric classification with efficient incremental learning via density-based support set selection.   
(ii) iNatIGCD, a new in-the-wild dataset for benchmarking IGCD that features a multi-stage training split motivated by a real-world fine-grained visual categorization use-case.  
(iii) A thorough evaluation on IGCD where we outperform recent methods.

\section{Related Work}
Here we review existing work in semi-supervised learning and both standard and incremental category discovery.

\noindent\textbf{Semi-Supervised Learning.}
Conventional semi-supervised learning assumes that we have access to both labeled and unlabeled data at training time~\cite{zhu2005semi,sslbook2006,oliver2018realistic}. 
Many works have been proposed to tackle this task using pseudo-labeling~\cite{rizve2021defense}, consistency regularization~\cite{berthelot2019mixmatch,tarvainen2017mean,sohn2020fixmatch,laine2016temporal}, density-based label propagation~\cite{li2020density}, or non-parametric categorization~\cite{assran2021semi}.
It is typically assumed that the unlabeled data contains instances from the \emph{same} categories that are present in the labeled set. 
Recent works have extended semi-supervised learning by removing the assumption that the categories in the unlabeled and labeled sets are the same~\cite{saito2021openmatch,huang2021trash,yu2020multi}, but their focus is still on the performance on the labeled categories and they do not evaluate the clustering accuracy on the novel categories in the unlabeled set.

\noindent\textbf{Open-set and Open-world recognition.}
Open-set recognition considers the case where novel classes can appear during testing, and the model needs to reject those novel classes~\cite{scheirer2012toward,geng2020recent,bendale2016towards}.
Thus open-set recognition method cannot be directly used for category discovery as them only rejects the novel categories.
Similarly, open-world recognition methods deal with the novel categories by incrementally soliciting human labels for the novel categories~\cite{bendale2015towards,boult2019learning,mundt2023wholistic,mundt2022unified}. 
While in category discovery, we require the models to automatically discover the novel categories without the human-in-the-loop.

\noindent\textbf{Category Discovery.} 
In contrast to semi-supervised learning, Novel Category Discovery (NCD)~\cite{han2019learning}  addresses an alternative setting whereby there is no overlap between the categories in the labeled and unlabeled sets. 
Here, the goal is to automatically discover the novel categories in the unlabeled data.  
This can be viewed as a semi-supervised clustering problem~\cite{hsu2018learning,hsu2019multi,han2019learning}. 

Recently, Generalized Category Discovery (GCD), a more realistic and challenging version of the discovery problem has been proposed~\cite{cao22,vaze2022generalized}. 
In this setting, the unlabeled data can consist of images from \emph{both} seen and unseen (\ie novel) categories.
Earlier works showed that self-supervised pretraining can aid category discovery~\cite{Han2020automatically}. 
Starting from a backbone initialized using self-supervised pretraining~\cite{chen2020simple}, ORCA~\cite{cao22} assumes the number of novel categories is known and proposed a three-component loss to train a deep classifier. 
The loss is comprised of a supervised component, a pairwise loss that uses high confident pseudo-labels to enforce that similar unlabeled instances are grouped together, and a regularization term to ensure that all instances are not assigned to the same category. 
GCD~\cite{vaze2022generalized} also make use of self-supervised pretraining using \cite{caron2021emerging}, but foregoes the need for pseudo-labels by instead using a clustering-based approach. 
They perform a contrastive finetuning step using image pairs from the same category for the labeled data and augmented pairs for the unlabeled data. 
Finally, they perform unsupervised clustering using $k$-means to assign unlabeled images to categories. 

Building on GCD~\cite{vaze2022generalized}, SimGCD~\cite{wen2023simple} investigated the impact of different design choices on downstream performance. 
Their final approach does not use an  explicit clustering step but instead makes use of learned category prototypes inspired by self-distillation methods~\cite{assran2022masked}, resulting in improved performance compared to GCD.  
MIB~\cite{chiaroni2022mutual} employs a similar pretraining phase to GCD  to train their feature extractor, %
in addition to a cross-entropy and conditional entropy loss for the labeled and unlabeled data respectively. 
We take inspiration from these methods, but explore a different setting of the discovery problem, that of IGCD.

\noindent\textbf{Incremental Category Discovery.} 
Category-incremental learning is a learning setting whereby the number of categories in a dataset increases over time, \eg over a set of discrete learning stages. 
The challenge in the incremental setting is that it is not possible to store all of the previously observed training data from earlier stages during each subsequent learning stage. 
Some of the difficulties that arise from this setting include catastrophic forgetting~\cite{mccloskey1989catastrophic,kirkpatrick2017overcoming} (\ie a dramatic decrease in performance on old categories as a result of training on new ones) and susceptibility to distributional shifts in the data. 
Multiple different approaches have been explored in the literature to address this, including storing past training examples in a replay buffer~\cite{robins1995catastrophic} or storing distilled exemplars~\cite{rebuffi2017icarl,buzzega2020dark}. Taxonomic variants of the problem have also been explored~\cite{lin2022continual,chen2023taxonomic}. 
For a recent survey on the topic we refer readers to~\cite{zhou2023deep}. 

Most relevant to our work are a recent set of approaches that explore category discovery in the incremental/continuous setting, \eg~\cite{roy2022class,joseph2022novel,zhang2022grow,liu2023large}.  
\cite{roy2022class} proposed a one-stage category-incremental setting termed class-iNCD. 
Here labeled data is initially available to learn a representation, then discarded, and finally a set of unlabeled data containing only novel categories is provided. 
Their goal is to train a model that performs well on both sets of categories. 
However, here there is no overlap between the categories in the labeled and unlabeled sets. 
This setting has also been explored in~\cite{liu2022residual}, where their focus was only on the category discovery performance. 	
FRoST~\cite{roy2022class} retains feature prototypes learned from  labeled data which are replayed during the discovery phase to prevent forgetting of the old categories. 
NCDwF~\cite{joseph2022novel} also explore the same setting and propose a method that uses pseudo-latent supervision, feature distillation, and a mutual information-based regularizer to maintain performance on the discarded labeled categories and assist in discovering novel ones. 
In the MSc-iNCD setting explored in~\cite{liu2023large}, the model only obtains unlabeled data at each time step which is assumed to come from previously unseen (\ie novel) categories. 

Grow and Merge (GM)~\cite{zhang2022grow} was recently introduced and applied to several different incremental discovery settings. 
It performs two key phases at each learning stage. 
First in the growing phases, it performs novelty detection to separate novel from seen categories and then trains a dynamic network to perform NCD.
Then in the merging phases, it combines the newly discovered categories with the previously know ones into a single model. 
Later, we compare to GM and show superior performance. 

With the exception of GM, who only perform one related experiment on CIFAR-100~\cite{cifar}, other existing methods assume that unlabeled data can \emph{only} contain novel categories. 
In contrast, we explore the more challenging incremental \emph{generalized}  setting where, at each stage, new data can come from \emph{either} previously seen or novel categories.
Unlike~\cite{zhang2022grow}, in our setting, all previously seen categories are not guaranteed to be present in each subsequent stage. 

\noindent\textbf{Benchmarking Category Discovery.} 
There are several common datasets used to evaluate category discovery methods. 
Earlier works~\cite{han2019learning,Han2020automatically} typically use standard image categorization datasets like CIFAR10/100~\cite{cifar} and ImageNet~\cite{imagenet} by creating artificial splits for discovery evaluation. 
\cite{zhao21novel} argue that more challenging fine-grained datasets like CUB~\cite{WahCUB_200_2011} are more suitable for evaluating discovery performance as the labeled and unlabeled categories share more visual similarity. 
Recently, the Semantic-Shift Benchmark (SSB)~\cite{vaze2022openset,vaze2022generalized} was proposed to better evaluate the task of detecting semantic novelty. 
SSB uses the existing CUB~\cite{WahCUB_200_2011}, Stanford Cars~\cite{cars}, and FGVC-Aircraft~\cite{fgvc_aircraft} datasets where the category splits are designed to have a clear `axes of semantic variation' as well as a coherent definition of categories in the labeled and unlabeled sets. 

Recent work in incremental discovery~\cite{roy2022class,joseph2022novel,zhang2022grow} evaluate performance using artificial data splits on CIFAR-10/100~\cite{roy2022class,zhang2022grow,joseph2022novel}, TinyImageNet~\cite{roy2022class}, ImageNet~\cite{joseph2022novel}, or the fine-grained SSB datasets~\cite{zhang2022grow}. 
However, these artificial data splits may not reflect the real-world performance of IGCD methods. 
For example, it has been shown that the category splits in the labeled dataset play an important role NCD performance~\cite{fei2022xcon}. 
To address this gap, we introduce a new `real-world' data split derived from the publicly available fine-grained iNaturalist species categorization dataset~\cite{van2021benchmarking}. 
We make it applicable to the incremental setting by leveraging metadata, such as the date and location of each observation. %
From this, we are able to better simulate real-world IGCD by creating data and category splits that mimic those that naturally arise over multiple time steps in the real-world. 
Note that the geographical domain shift has also been studied in GeoNet~\cite{kalluri2023gnet}, where the main task is unsupervised domain adaptation. 
Besides the task difference, our iNatIGCD dataset also contains $10\times$ more images, from more geographical locations, and fine-grained concepts that are challenging for category discovery.

\section{The iNatIGCD Benchmark}
Here we outline iNatIGCD, our new benchmark for IGCD.  
Our dataset is based on the large-scale fine-grained iNat2021 visual categorization dataset~\cite{van2021benchmarking}, which contains images from 10,000 different categories of plant and animal species sourced from the community science platform iNaturalist~\cite{iNatWeb}. 
Each of the images posted to iNaturalist contains the precise capture date and time, along with the geographical location as metadata~\cite{iNatWeb}. 
This creates a natural stream of data over time that contains category distribution shifts between observations from around the globe as the iNaturalist community grows across different regions.

iNatIGCD is motivated by the scenario where one can develop an initial image classifier from the currently available labeled data, which only contains images and labels captured before a certain time and is restricted to a certain region (\eg a specific continent). 
As time progresses, new unlabeled data is obtained which can also include images from other regions (\eg a different continent). 
This scenario is close to the real-world situation of the iNaturalist platform where we  have access to labeled image observations at a given time point, but we cannot easily obtain reliable labels for new observations that are shared each day as labeling images takes time. 
We can view the existing collection of observations as the labeled dataset and the new observations, which can contain a combination of old and new categories, as the unlabeled dataset. 
Similar to the common assumption in continual learning~\cite{rebuffi2017icarl,buzzega2020dark}, all the previous labeled data can be difficult, or even impossible, to store during learning. 
Thus as time progresses, we assume that we do \emph{not} have access to all the past labeled data when  retraining/updating the model.

Our new  IGCD benchmark is structured as follows:
At each time step, we have a labeled dataset and an unlabeled one. 
Our goal is to correctly discover any novel categories in the unlabeled dataset while  also classifying examples belonging to known categories. 
When we progress to a new time step, the labeled dataset from the previous step becomes unavailable. 
However, we now have a new labeled dataset, which is actually the previous unlabeled dataset, along with a new unlabeled dataset which is used for category discovery. 
This simulates the real-world process as it takes time for the iNaturalist community to reach a consensus regarding which category is present in a newly uploaded image.  
See \cref{fig:teaser} for an illustration of the process.

\begin{figure}[t]
\centering
    \includegraphics[width=1.0\columnwidth]{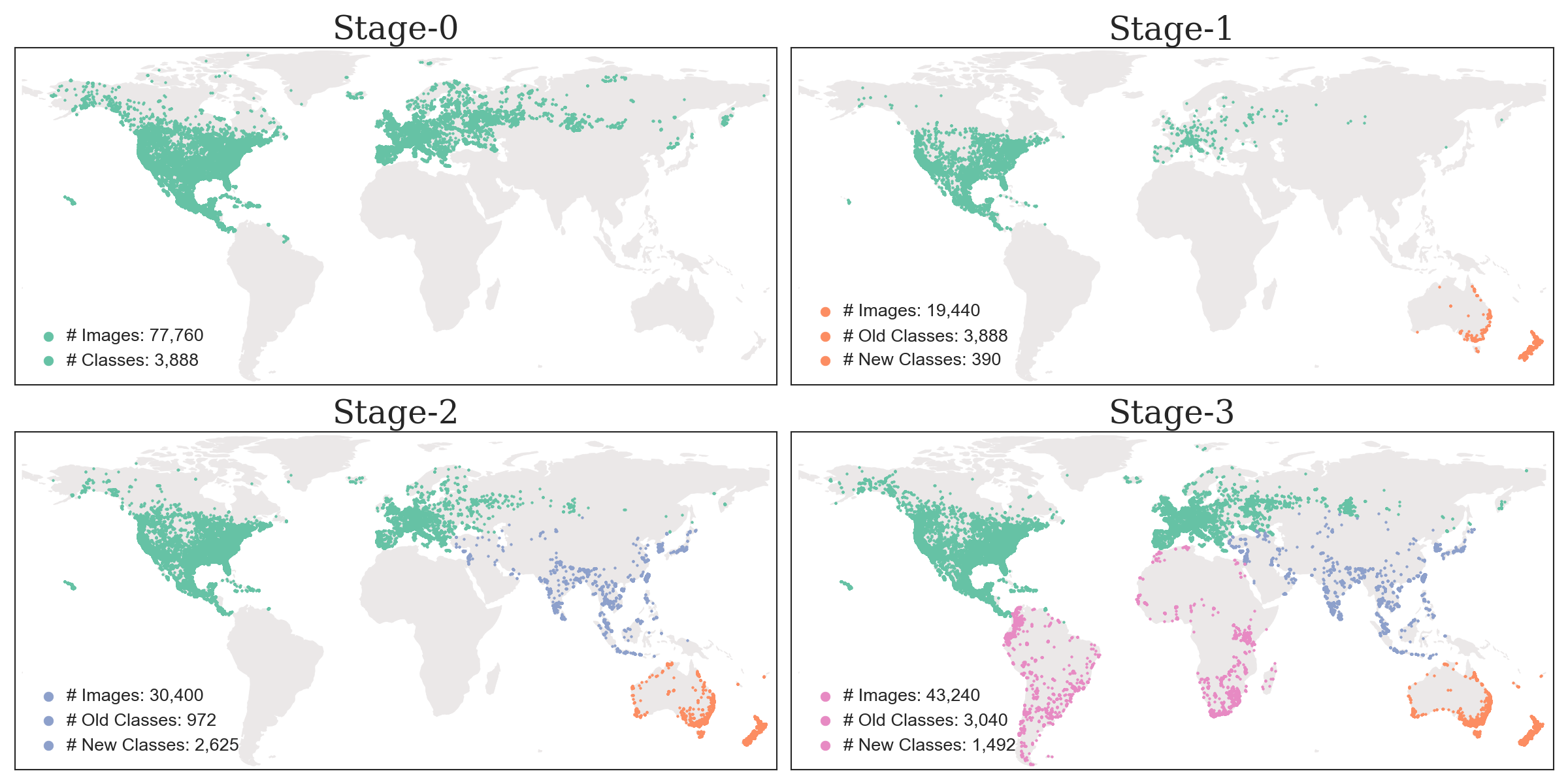}
\vspace{-20pt}
\caption{
A visualization of the location of the data in each stage of our iNatIGCD dataset. We also report the number of old and new categories for each of the stages.
}
\label{fig:world_map}
\vspace{-10pt}
\end{figure}

\noindent{\bf Dataset Construction.}  To generate our data splits, we leverage metadata from the iNat2021 dataset~\cite{van2021benchmarking}. 
We first sort and group all the images according to the time when they were captured. 
To make the data more evenly distributed, we cluster them into four temporal stages: \mbox{2008-2016}, 2017, 2018, and 2019. 
We further split the data according to the location of the observation, only including data from North America and Europe in the first stage and then including Oceania, Asia, and finally Africa and South America in the second, third, and fourth stages respectively. 
This simulates the growth of the species recognition community on iNaturalist from continent to continent over time. 

Our new iNatIGCD dataset is summarized in~\cref{fig:world_map}.  
It poses several unique challenges for category discovery methods: (i) we have an order of magnitude more finer-grained categories in each of the incremental stages compared to existing datasets which will challenge the ability of category discovery methods, (ii) at each stage, we have more images that require categorization, which can give a more reliable measure of the discovery performance of a model, and (iii) the categories that appear in each stage are based on the natural distribution of species rather than an artificial split as in previous benchmarks. 

\begin{figure*}[t]
\centering
    \includegraphics[width=0.9\linewidth]{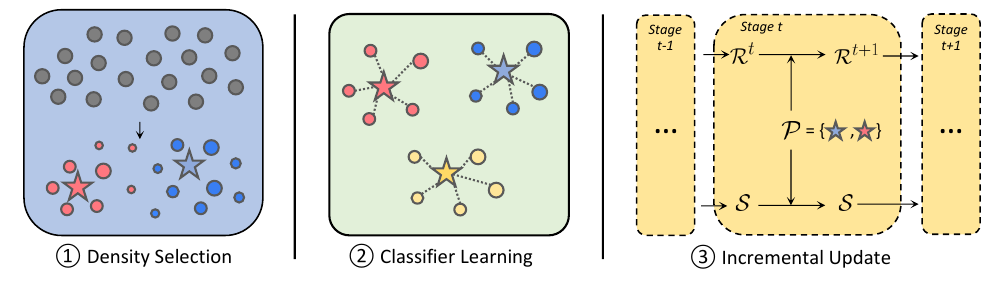}
\vspace{-10pt}
\caption{
Overview of our method. (1) Illustration of the density calculation process. 
We denote the density of each data point by its size in the lower part. 
The density peaks, \ie the data point whose density is higher than its neighbors, are illustrated with a star. 
(2) During classifier learning we update representations and classifiers using the loss defined in~\cref{eq:overall}.
(3) The incremental update procedure for stage $t$. 
When proceeding to stage $t+1$, the density peaks $\mathcal{P}$ are added to $\mathcal{R}^{t+1}$ and $\mathcal{S}$.
}
\label{fig:method}
\vspace{-10pt}
\end{figure*}

\section{Method}

In this section we present a new approach targeting the IGCD setting.
At each stage~$t$ (\ie time step), we have two datasets to train the model on, $\mathcal{D}_{lab}^t$ and $\mathcal{D}_{unlab}^t$. 
Our goal is to learn a classifier that can correctly classify and discover categories in  $\mathcal{D}_{lab}^t$ and $\mathcal{D}_{unlab}^t$, in addition to  maintaining its performance on the categories in the current stage when we advance to later stages. 

In the IGCD setting, it is assumed that it is not possible to store all the previously observed data. 
As a result, the model must learn incrementally. 
A common incremental learning technique is to avail of the concept of example replay~\cite{rebuffi2017icarl,buzzega2020dark}. 
This involves saving a few examples of each category and then `replaying' them to the model during training in later stages. 
However, there are challenges associated with applying replay approaches in our IGCD setting due to two primary issues: (i) we have novel categories in the unlabeled dataset in IGCD, thus it is not straightforward to select examples for these novel categories as previous works typically assume full supervision, and (ii) SoTA GCD methods use parametric classifiers~\cite{wen2023simple,uno}, which we show in experiments can overfit to the few replay examples. 
We tackle these technical challenges by using a non-parametric classifier combined with a density-based example selection mechanism which we introduce below.

\subsection{Problem Setting} 
Formally, we  denote the labeled dataset at the initial stage as $\mathcal{D}_{lab}^{0}=\{(\bm{x}_i^0, \bm{y}_i^0)\}$. 
This is used to train an initial model to recognize the set of categories $\mathcal{C}^0=\{1,2,\dots,K^0\}$. 
For incremental learning at a later stage $t$, we have two datasets,  a labeled dataset $\mathcal{D}_{lab}^t=\{(\bm{x}_i^t,\bm{y}_i^t)\}$ and an unlabeled dataset $\mathcal{D}_{unlab}^t=\{\bm{x}_i^t\}$. 
We denote the category sets in $\mathcal{D}_{lab}^t$ and $\mathcal{D}_{unlab}^t$ as $\mathcal{C}_{lab}^t$ and $\mathcal{C}_{unlab}^t$. 
In our generalized setting, $\mathcal{C}_{lab}^t$ and $\mathcal{C}_{unlab}^t$ can be different and may only have partial overlap.
When we progresses to stage $t+1$, $\mathcal{D}_{unlab}^t$ will receive its labels which are used as $\mathcal{D}_{lab}^{t+1}$ for the next stage.
The goal is to learn a model that can not only discover novel categories in the unlabeled dataset $\mathcal{D}_{unlab}^t$ at stage $t$, but also maintain the ability to recognize the categories in $\mathcal{C}_{lab}^0,\mathcal{C}_{lab}^1,\dots,\mathcal{C}_{lab}^{t}$.
Similar to the category-incremental learning setting~\cite{rebuffi2017icarl,buzzega2020dark}, one main added challenge of our case, compared to static GCD, is that the model can exhibit catastrophic forgetting~\cite{robins1995catastrophic} on the categories that are not present in $\mathcal{C}_{lab}^t$ and $\mathcal{C}_{unlab}^t$. 

Note, our setting is related, but different from the Mixed Incremental (MI) one in~\cite{zhang2022grow}. 
In contrast to us, \cite{zhang2022grow} assume that only the unlabeled data $\mathcal{D}_{unlab}^t$ at stage $t$ is available for training the model. 
In our experiments, we also compare models that are only trained with $\mathcal{D}_{unlab}^t$. %
Furthermore, in the MI setting, the category set $\mathcal{C}_{unlab}^t$ always includes \emph{all} previously seen categories. 
In our more difficult and realistic setting, the category set at each time step may only have partial overlap with the previous category sets.

\subsection{Classifier Learning}
\label{sec:classifier}
For each input image $\bm{x}_i$, we use a neural network $f$ to extract its corresponding feature representation $\bm{h}_i=f(\bm{x}_i)$. 
From this, we use a non-parametric classifier $g$ to assign the image to a category probability vector $\bm{p}_i=g(\bm{h}_i)$, where $\sum_c \bm{p}_{i,c} = 1$. %
For $f$ to learn an informative representation, we adopt a supervised contrastive loss $\mathcal{L}_{\text{SupCon}}$ from~\cite{khosla2020supervised} for learning from $\mathcal{D}_{lab}$ and use a self-supervised contrastive loss $\mathcal{L}_{\text{SelfCon}}$ from~\cite{chen2020simple} for learning from $\mathcal{D}_{unlab}$. 
These two losses are combined via a balancing factor $\lambda_{\text{rep}}$, 
\begin{equation}
\mathcal{L}_{\text{rep}}=\lambda_{\text{rep}} \mathcal{L}_{\text{SupCon}} + (1-\lambda_{\text{rep}}) \mathcal{L}_{\text{SelfCon}}.
\end{equation}

We use a non-parametric Soft Nearest-neighbor (SNN) classifier for $g$ that computes the distribution over category labels for an input image $\bm{x}_i$, 
\begin{equation}
    \bm{p}_i=\texttt{SNN}(\bm{h}_i,\mathcal{S},\tau)=\sum_{k=1}^{N^\mathcal{S}} \frac{\exp (\bm{h}_i \cdot \bm{h}^l_k / \tau)}{\sum_{\bm{h}^{l}_b\in \mathcal{S}} \exp (\bm{h}_i \cdot \bm{h}^{l}_{b} / \tau)} \bm{y}_k.
\end{equation}
$\mathcal{S}=\{(\bm{h}^l_k, \bm{y}_k)\}$ is a support set containing features of all support samples $\bm{h}^l_k$ and their corresponding one-hot labels $\bm{y}_k$,  $N^{\mathcal{S}}=|\mathcal{S}|$ is the number of support samples in $\mathcal{S}$, and $K^{\mathcal{S}}$ is the number of categories in $\mathcal{S}$. 
$\bm{h}_i$ is the representation of $\bm{x}_i$ from $f$, and $\tau$ is a temperature parameter controlling the sharpness of the prediction $\bm{p}_i$. 
The support samples are representative examples of each of the categories and are used to softly assign a label to the input example $\bm{h}_i$ by averaging all the one-hot labels $\bm{y}_k$ in $\mathcal{S}$ weighted by the softmax similarity. 
For the labeled categories, we randomly select the support examples. 
Later we discuss the selection of support examples for the novel categories. %

Importantly, as our classifier is non-parametric, it is easy to extend it so that it can classify more categories by simply adding additional examples from the new categories to the support set. 
At test time, we just use the saved support examples to perform categorization for an input image by assigning the labels via measuring the distance between the input image and support examples.
The classifier is trained with two cross-entropy losses $\mathcal{L}_{\text{ce}}$, and makes use of an entropy-regularizer,
\begin{align}
    \mathcal{L}_{l}&=\frac{1}{|\mathcal{B}^l|}\sum_{i\in \mathcal{B}^l}\mathcal{L}_{\text{ce}}(\bm{y}_i^l,\hat{\bm{p}}_i^l),  \\
    \mathcal{L}_{u}&=\frac{1}{|\mathcal{B}^u|}\sum_{i\in \mathcal{B}^u}\mathcal{L}_{\text{ce}}(\tilde{\bm{p}}_i^u, \hat{\bm{p}}_i^u) - \epsilon H(\overline{\bm{p}}),
\end{align}
where $\mathcal{B}^l$ and $\mathcal{B}^u$ are batches of indices from the labeled and unlabeled datasets, 
$\bm{y}_i^l$ is the ground-truth label for the example $\bm{x}_i^l$, and $\tilde{\bm{p}}_i^u$ and $\hat{\bm{p}}_i^u$ are the predictions for two augmented views of the same input example $\bm{x}_i^u$ with different temperatures. 
These two losses together form the loss function we use for classifier learning,
\begin{equation}
\mathcal{L}_{\text{cls}}=\lambda_{\text{cls}}\mathcal{L}_{l} + (1-\lambda_{\text{cls}})\mathcal{L}_{u},
\end{equation}
where $\lambda_{\text{cls}}$ is a weighting factor.
The entropy regularization term, $H(\overline{\bm{p}})=-\sum \overline{\bm{p}} \log \overline{\bm{p}}$, regularizes the mean prediction $\overline{\bm{p}}$ over a mini-batch which is computed as 
\begin{equation}
\overline{\bm{p}}=\frac{1}{2|\mathcal{B}^u|}\sum_{i\in \mathcal{B}^u} (\tilde{\bm{p}}_i^u + \hat{\bm{p}}_i^u).
\end{equation} 
This above entropy term has been adopted in previous works in GCD~\cite{wen2023simple} and semi-supervised learning~\cite{assran2021semi} to calibrate predicted category distribution and to avoid empty clusters, as observed in~\cite{vaze2022generalized,wen2023simple}.
The overall loss for training the model is
\begin{equation}\label{eq:overall}
\mathcal{L}=\mathcal{L}_{\text{rep}}+\mathcal{L}_{\text{cls}}. 
\end{equation}

\subsection{Support Sample Selection} 
\label{sec:subport_samp}
Our $\texttt{SNN}$ classifier function is similar to the one defined in the semi-supervised approach of~\cite{assran2021semi}. 
However, in its original form, it cannot be used to learn from novel categories from the unlabeled data if they do not already appear in the labeled set. 
We extend this approach via a novel modification to make it applicable to the generalized category discovery setting.
Specifically, for the potentially novel categories in the unlabeled dataset, we propose a new density-based selection mechanism to select a subset of examples from the unlabeled dataset and then pseudo-label them with $g$ to use them as support samples in~$S$. 

We start by estimating the density $d_i$ of a sample $\bm{x}_i^u$ based on its $k$-nearest neighbors, such that
\begin{equation}
    d_i=\frac{1}{K} \sum_{j\in \mathcal{N}_{\bm{x}_i^u}}^K \frac{\bm{h}_i^u \cdot \bm{h}_j^u}{\|\bm{h}_i^u\|_2 \|\bm{h}_j^u\|_2}. 
\end{equation}
Here we iterate through the top-$K$ neighbors and average the cosine similarities of the feature representations, where 
$\mathcal{N}_{\bm{x}_i^u}$ contains the indices of the $K$ nearest neighbors of $\bm{x}_i^u$. 
Intuitively, a larger density value $d_i$ indicates that the corresponding prediction for $\bm{x}_i^u$ is a more reliable pseudo-label as it indicates that $\bm{x}_i^u$ is similar to its neighbors in the learned representation space of $f$, see~\cite{li2020density}.
Inspired by this, we propose a method for automatically selecting reliable samples from $\mathcal{D}_{unlab}$ using this density definition.

First, we select the density peaks from the dataset which are the points whose density is higher than all other $k$-nearest neighbors around it~\cite{li2020density}. 
Specifically, the set of density peaks can be defined as
\begin{equation}
    \mathcal{P} = \{\bm{x}_i^u | \forall k \in \mathcal{N}_{\bm{x}_i^u}^K, d_i > d_k\},
\end{equation}
where $\mathcal{P}$ is a set of selected images $\bm{x}_i$.
Second, it is not guaranteed that there is only one density peak per class. 
To address this, we use an intersection-over-union (IoU) score on the nearest neighbor set between two density peaks to measure their similarity. 
We then remove redundant peaks using a similar procedure to non-maximum-suppression~\cite{nms}. 

Thus, for two density peaks $\bm{x}_i^u$ and $\bm{x}_j^u$ in $\mathcal{P}$, we  measure their similarity by calculating the IoU score of their $K^d$ nearest neighbor set $\mathcal{N}_{\bm{x}_i^u}^{K^d}$ and $\mathcal{N}_{\bm{x}_j^u}^{K^d}$. 
If the IoU score is higher than a threshold $T$, we only keep the peak with the higher density value. 
The selected density peaks are used to pseudo-label their close neighbors based on distance, and then the density peaks and their closer neighbors are added as support samples for novel categories to $\mathcal{S}$. 
For the labeled categories, we select a number of support examples from the labeled dataset using the same density selection mechanism to serve as their support samples. 
    This procedure is outlined in detail in the supplementary material.

The benefit of our proposed method is that we automatically have an estimate of the number of novel categories in the unlabeled set based on the number of selected density peaks. 
This is in contrast to previous methods for estimating the number of categories which make use of the $k$-means algorithm, \eg~\cite{han2019learning,vaze2022generalized}. 
Our approach is more closely integrated with the representation learning step and is also more computationally efficient.

\subsection{Incremental Update}
\label{sec:incremental}
Earlier we introduced our method for learning a classifier that leverages non-parametric category assignments. 
Now we discuss how to extend this model to the \emph{incremental} setting. 
To do this, we focus on learning between a pair of stages, \ie from stage $t$ to $t+1$. 

After training on $\mathcal{D}_{lab}^t$ and $\mathcal{D}_{unlab}^t$ with the objectives defined in~\cref{sec:classifier}, we have a model that can classify the labeled and unlabeled categories, $\mathcal{C}_{lab}^t$ and $\mathcal{C}_{unlab}^t$. 
To handle subsequent stages, we draw inspiration from the classic category-incremental learning method iCaRL~\cite{rebuffi2017icarl}. 
Specifically, we store a set of representative examples $(\bm{x}_i, \bm{y}_i)$ to a memory buffer $\mathcal{R}^t$ for replay to be used for training at later stages. 
Note, this is not to be confused with the support set $S$ used by our non-parametric classifier defined earlier. 

Different from iCaRL which selects examples using category centroids, we use our proposed density selection mechanism described in~\cref{sec:subport_samp} to select the examples to save to $\mathcal{R}^t$.
Thus at stage $t+1$, we concatenate the training set  $\mathcal{D}_{lab}^{t+1}$ with the memory buffer $\mathcal{R}^t$ and then optimize the training objectives in~\cref{sec:classifier}. 
The memory buffer created at stage $t+1$ is also  concatenated with $\mathcal{R}^t$ to ensure that the knowledge from all previous stages is maintained.

\begin{table*}[t]
\centering
\resizebox{0.95\textwidth}{!}{
\begin{tabular}{l | c | ccc | cccc | ccccc | cc}
\toprule
Methods                                 & Stage-0      & \multicolumn{3}{|c}{Stage-1}  & \multicolumn{4}{|c}{Stage-2}   & \multicolumn{5}{|c}{Stage-3}    &  \multicolumn{2}{|c}{Overall}    \\ 
\midrule
                                         & All          & All  & Old  & New           & All  & Old  & New    & S-0    & All  & Old  & New   & S-1  & S-0   & $M_f\downarrow$     &  $M_d$  \\
\midrule
Supervised upper-bound                   & 42.3         & 62.8 & 68.7 & 59.8          & 65.8 & 71.5 & 62.3   & 48.2   & 68.5 & 72.5 & 64.3  & 58.4 & 49.7  & - & - \\ 
\midrule
SimGCD~\cite{wen2023simple}              & 42.3            & \uline{24.1} & \textbf{35.8} & 10.5          & \uline{25.6} & \textbf{36.8} & \uline{18.4}   & 2.5    & 22.4 & \textbf{36.7} & 16.4  &  5.4 &  1.3  & 41.0  & 24.5   \\
SimGCD + iCaRL~\cite{rebuffi2017icarl}   & 41.5         & 23.7 & 33.6 & \uline{11.9}          & 23.5 & 34.8 & 12.7   & 9.4    & \textbf{23.5} & 35.8 & \uline{17.5}  & 11.3 & 8.6  & 32.9 & \uline{27.9}    \\
FRoST~\cite{roy2022class}                & 42.3            & 20.5 & 24.6 & 9.4           & 12.4 & 24.1 & 7.8    & 7.3    & 13.5  &  18.5 & 7.6   & 6.5  &  4.2  & 38.1 & 19.7 \\
GM~\cite{zhang2022grow}                  & 42.3            & 18.7 & 26.7 & 8.6           & 19.5 & 28.7 & 10.4   & \uline{14.0}   & 16.5 & 25.8 & 13.5  & \textbf{16.8} & \uline{12.3} & \uline{30.0} & 20.1 \\
\midrule
Ours                                     & 41.8         & \textbf{25.6} & \uline{34.6} & \textbf{14.5}          & \textbf{25.7} & \uline{35.0} & \textbf{21.5}   & \textbf{16.4}   & \uline{22.8} & \uline{35.9} & \textbf{17.8}  & \uline{16.4} & \textbf{14.2} & \textbf{27.6} & \textbf{28.4}    \\
\bottomrule
\end{tabular}
}
\vspace{-5pt}
\caption{Results on iNatIGCD in the IGCD-l setting (\ie where labels are available at the end of each stage). Higher numbers are better, with the exception of $M_f$ where lower is better.
}
\vspace{-15pt}
\label{tab:main_result_w_label}
\end{table*}

\begin{figure*}[h]
    \centering
    \subfloat{
        \includegraphics[width=0.4\textwidth]{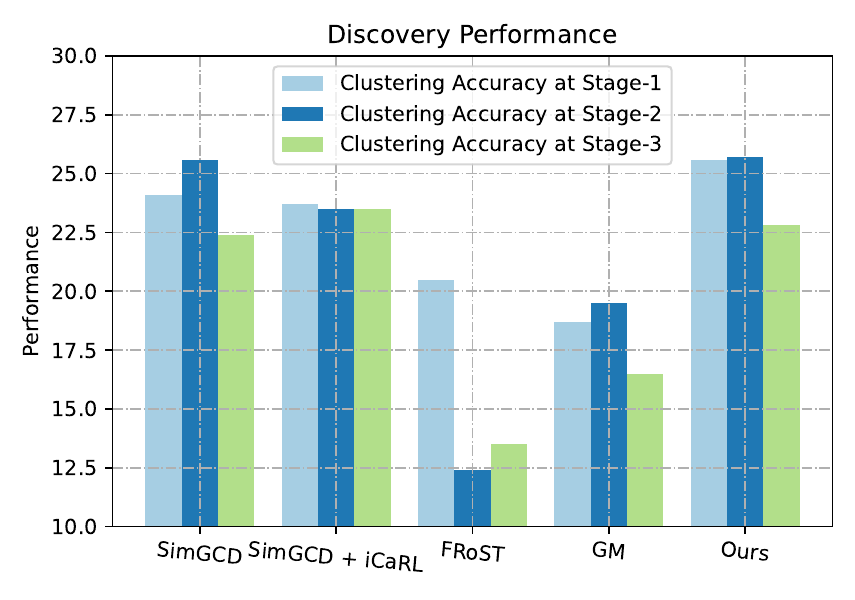}
        \label{fig:discovery_igcd_l}
    }
    \subfloat{
        \includegraphics[width=0.4\textwidth]{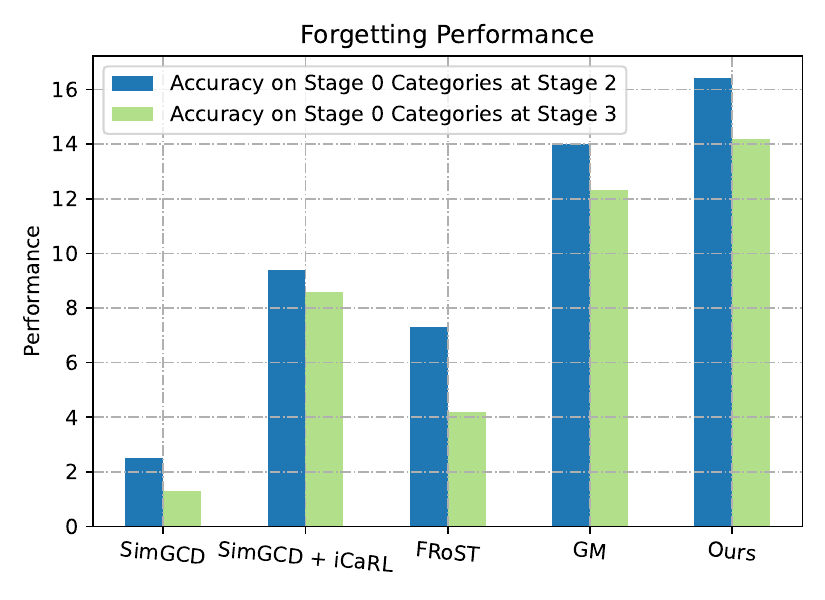}
        \label{fig:forgetting_igcd_l}
    }
    \vspace{-15pt}
    \caption{
    The category discovery performance (measured by `All' class clustering accuracy) and the forgetting (measured by accuracy on stage-0 classes) at each stage of the training. We can see that our proposed method obtains strong discovery performance, similar to SimGCD~\cite{wen2023simple}, and at the same time exhibits less forgetting.
    }
    \label{fig:discovery_forgetting}
    \vspace{-10pt}
\end{figure*}

\section{Experiments}

\subsection{Implementation Details}

\noindent{\bf Model Training.} We use a ResNet-18~\cite{he2016deep} initialized using MoCo~\cite{he2019moco} on ImageNet-1k as our feature extractor $f$. 
We freeze its parameters up to the last residual block to prevent the model from overfitting on the labeled data as in~\cite{Han2020automatically,zhao21novel}. 
We train for $100$ epochs on the labeled data from the initial stage $\mathcal{D}_{lab}^0$, and then train on each subsequent stage $t$ with $\mathcal{D}_{lab}^t$ and $\mathcal{D}_{unlab}^t$ for $40$ epochs each. 
We provide additional vision transformer~\cite{dosovitskiy2020vit} results and  details of the training settings in the supplementary material.

\noindent{\bf Evaluation Metrics.} 
We use our iNatIGCD dataset to evaluate the performance of different IGCD methods, using two different settings: (i) IGCD-l, where we assume the labels of $\mathcal{D}_{unlab}^t$ will become available at the end of stage $t$ and thus can be used as $\mathcal{D}_{lab}^{t+1}$, and (ii) IGCD-u, where we assume no data is labeled in any of the incremental stages, and at stage $t$ only $\mathcal{D}_{unlab}^{t}$ is available to train the model. 
We remove the supervised losses accordingly in IGCD-u.

For our experiments on iNatIGCD, we report the clustering accuracy for categories in $\mathcal{C}_{lab}^t$ and $\mathcal{C}_{unlab}^t$ at stage $t$ as `Old' and `New', and `All' as the performance on both. 
We also report the accuracy on the categories in previous stages $0,1,\dots,t-1$ that do not appear in the current stage as `S-$t$'.
As in~\cite{zhang2022grow}, we report maximum forgetting $\mathcal{M}_f$, which is the maximum difference between the clustering accuracy from the stage $0$ categories and any later stage $t$. 
We also report final discovery $\mathcal{M}_d$, which is the final clustering accuracy of the model on all the categories from all stages.

\subsection{Results}

Here we describe the results of relevant methods on different IGCD benchmarks.
We include comparisons to GM~\cite{zhang2022grow}, a recent IGCD method. 
We also compare to FRoST~\cite{roy2022class} which is a one-stage method that can only handle one incremental stage. 
We extend it to the multi-stage setting by repeating the second discovery phase of FRoST~\cite{roy2022class} for each new stage. 
We also compare to SimGCD~\cite{wen2023simple}, a recent SoTA GCD method and apply it to IGCD by rerunning it on the data from each new stage.
To enable it to maintain performance on previous categories, we combine it with a memory buffer as in iCaRL~\cite{rebuffi2017icarl}. 
Finally, to establish a performance upper bound on iNatIGCD, we train a fully supervised  model on all the data and labels at each stage $t$ (including all previous stages $t-1,\dots,0$). 
Additional results are in the supplementary material. 

\subsubsection{iNatIGCD Results}

\begin{table*}[h]
\centering
\resizebox{0.85\textwidth}{!}{
\begin{tabular}{l | c | cc | ccc | cccc | cc}
\toprule
Methods                                & Stage-0      & \multicolumn{2}{|c}{Stage-1}  & \multicolumn{3}{|c}{Stage-2}   & \multicolumn{4}{|c}{Stage-3}    & \multicolumn{2}{|c}{Overall}    \\ 
\midrule
                                       & All          & New & S-0                    & New & S-1 & S-0        &  New &  S-2 & S-1 & S-0  & $M_f\downarrow$   &  $M_d$  \\
\midrule
Supervised upper-bound                 & 42.3         & 59.8 & 48.7           & 62.3 & 63.9 & 49.6     &  64.3 & 64.3 & 64.1  & 52.4  & -  & - \\ 
\midrule
SimGCD~\cite{wen2023simple}            & 42.3            & 8.4 & 5.1            & 12.6 & 2.3 & 0.1      & 11.8 & 4.2 & 1.2  & 0.0  & 42.2  & 2.5 \\
SimGCD + iCaRL~\cite{rebuffi2017icarl} & 41.5         & \uline{10.5} &  16.7        & 13.4 & 8.4 & 11.2      & \uline{14.5} & 10.2 & 7.1  &  10.2 & 31.3  & \uline{12.4} \\
FRoST~\cite{roy2022class}              & 42.3            & 6.7 & 12.3           & 9.2 & 7.4 & 9.3       & 10.0 & 8.4 & 7.1  & 5.1  & 37.2  & 7.3 \\
GM~\cite{zhang2022grow}                & 42.3            & 8.6 & \uline{18.4}           & \uline{13.8} & \uline{9.1} & \textbf{15.5}     & 10.7  & \uline{11.0} & \uline{8.2}  & \uline{13.0}  & \uline{29.3}  & 10.5 \\
\midrule
Ours                                   & 41.8         & \textbf{12.7} & \textbf{22.1}           & \textbf{15.4} & \textbf{10.7} & \uline{14.3}      & \textbf{16.1} & \textbf{11.2} & \textbf{9.1}  & \textbf{13.1} & \textbf{28.7} & \textbf{14.1} \\
\bottomrule
\end{tabular}
}
\vspace{-5pt}
\caption{
Results on iNatIGCD in the IGCD-u setting (\ie where labeled data is not provided during the incremental stages). 
Higher numbers are better, with the exception of $M_f$ where lower is better. 
}
\vspace{-10pt}
\label{tab:main_result_wo_label}
\end{table*}

\noindent{\bf IGCD-l Setting.}
We first present results for the incremental setting with labels in~\cref{tab:main_result_w_label} and~\cref{fig:discovery_forgetting}. 
Although the non-incremental SimGCD~\cite{wen2023simple} achieves comparable performance at recognizing labeled categories ('Old') and discovering novel categories ('New') at each stage, it fails to maintain good performance on the categories it has seen before but that are not present in the current stage (see `S-0' and `S-1'). 
Combining SimGCD with iCaRL~\cite{rebuffi2017icarl} helps alleviate this forgetting problem, which we see as an increase in `S-0' and `S-1' scores compared to SimGCD alone. 
However, the performance on `Old' categories, which have labeled data during training, drops for SimGCD + iCaRL. 
We argue that this is because SimGCD employs a parametric classifier, but iCaRL~\cite{rebuffi2017icarl} performs non-parametric categorization at test time. 
We can see from the Stage-0 performance of SimGCD + iCaRL that this mismatch between training and testing results in a performance drop even in the initial fully supervised scenario. 
In contrast, our $\texttt{SNN}$ classifier achieves a balance between preventing the category information from previous stages from being forgotten (`S-0' and `S-1') as well as obtaining good performance on the categories it is currently trained on (`Old' and `New').

Compared to GM~\cite{zhang2022grow}, which is designed to tackle  discovery and forgetting simultaneously, our proposed method shows a clear advantage, with improved discovery performance (see $\mathcal{M}_d$) and reduced forgetting (see $\mathcal{M}_f$). 
We also outperform the recent FRoST~\cite{roy2022class} method. 
Notably, we always perform best on the novel categories at each stage (see `New'). 
We speculate that this is likely because GM~\cite{zhang2022grow} and FRoST~\cite{roy2022class}  employ a pair-wise objective for learning to cluster novel categories. 
However in iNatIGCD, the number of categories is high, thus there will be far more negative pairs than positive ones to train their methods, resulting in degraded performance.

\noindent{\bf IGCD-u Setting.}
In~\cref{tab:main_result_wo_label} we present results for the incremental setting where  $\mathcal{D}_{lab}^t$ is not available at each stage~$t$. 
Thus we report performance on `New' categories in $\mathcal{D}_{unlab}^t$ at each stage $t$ to evaluate a model's ability to discover new categories. 
We also report the performance on S-0, S-1, and S-2 to evaluate the forgetting of previously learned categories. 
Due to the lack of supervision from $\mathcal{D}_{lab}^t$, the performance of all models is reduced, yet our proposed approach achieves superior results in almost all cases.

\subsubsection{Mixed Incremental Results}
In~\cref{tab:MI} we present results on the Mixed Incremental (MI) scenario proposed in~\cite{zhang2022grow}. %
At each incremental stage $t$, the model is trained on a dataset $\mathcal{D}_{unlab}^t$ of unlabeled examples which contains \emph{both} novel categories and the categories the model has already learned before. 
The goal is to classify both novel and seen categories at the same time. 
The difference between MI and our IGCD-u setting is that in MI it is assumed that at each stage there are always instances from the previous stage's categories in the unlabeled images. 

GM~\cite{zhang2022grow} only report results for the MI setting on CIFAR-100. 
For fairness, we re-train their model using the same hyper-parameters as ours and observe improved performance for their method compared to their paper. 
We also present additional results on CUB~\cite{WahCUB_200_2011}.
We include comparisons to the strong GCD baseline SimGCD~\cite{wen2023simple}, which cannot learn incrementally, and our incremental extension of their method. 
We observe that SimGCD~\cite{wen2023simple} exhibits catastrophic forgetting, but adding iCaRL~\cite{rebuffi2017icarl} can alleviate this issue. 
Our proposed method achieves better performance than GM~\cite{zhang2022grow}, with the exception of the $M_f$ performance on the smaller CIFAR-100 dataset. 
\vspace{-5pt}

\begin{table}[h]
\centering
\resizebox{0.9\columnwidth}{!}{
\begin{tabular}{l |rr | rr}
\toprule
Methods        & \multicolumn{2}{c}{CIFAR-100} & \multicolumn{2}{|c}{CUB} \\
  & \multicolumn{1}{c}{$M_f \downarrow$} & \multicolumn{1}{c}{$M_d \uparrow$} & \multicolumn{1}{|c}{$M_f \downarrow$} & \multicolumn{1}{c}{$M_d \uparrow$} \\
\midrule
SimGCD~\cite{wen2023simple}            & 58.7          & 28.3          & 63.5       & 27.4       \\
SimGCD + iCaRL~\cite{rebuffi2017icarl} & 9.4           & 29.4          & 10.7       & \uline{28.3}       \\
GM~\cite{zhang2022grow}                & \textbf{3.6}  & \uline{30.6}  & \uline{6.8}        & 26.7       \\
\midrule
Ours                                   & \uline{4.0}   & \textbf{31.2} & \textbf{6.7}        & \textbf{29.4}      \\
\bottomrule
\end{tabular}
}
\vspace{-5pt}
\caption{Results on the MI setting introduced in~\cite{zhang2022grow}. 
}
\vspace{-10pt}
\label{tab:MI}
\end{table}

\subsubsection{Ablations}

\noindent{\bf Estimating the Number of Novel Categories.} 
One of the unique challenges in GCD is the task of estimating the number of novel categories in the unlabeled data.
\cite{vaze2022generalized} described a baseline that uses semi-supervised $k$-means to estimate it. 
GM~\cite{zhang2022grow} adopted this method to estimate the number at each incremental stage before training their classifier.
Our proposed density selection method can also be seen as a way of estimating the number of categories based on the number of density peaks. 
In~\cref{tab:k_est}, we demonstrate that our method is capable of providing a more accurate estimate  compared to GM, without requiring their expensive $k$-means step. 
Note, the ground truth class count for iNatGCD-l in~\cref{tab:k_est} indicates the number of all categories including `Old' and `New', thus it is different from the counts in~\cref{fig:world_map}. 

\begin{table}[h]
\centering
\resizebox{0.80\columnwidth}{!}{
\begin{tabular}{lccc}
\toprule
CIFAR-100 MI             & \# Classes ($t=1$) & $t=2$ & $t=3$     \\
\midrule
Ground truth             & 10 & 10 & 10    \\
GM~\cite{zhang2022grow}  & 14 & 13 & 13    \\
Ours                     & 13 & 10 & 12    \\
\midrule
iNatIGCD-l               & \# Classes ($t=1$) & $t=2$ & $t=3$      \\                 
\midrule
Ground truth             & 972  & 3,040 & 4,324 \\
GM~\cite{zhang2022grow}  & 857  & 2,563 & 3,854 \\
Ours                     & 886  & 2,857 & 4,085 \\
\bottomrule
\end{tabular}
}
\vspace{-5pt}
\caption{
Estimation of the number of novel categories in the unlabeled data at each stage.
}
\label{tab:k_est}
\end{table}

\noindent{\bf Impact of the Selection Method for $\mathcal{R}$.}
Compared to the centroid selection in iCaRL~\cite{rebuffi2017icarl}, our method uses the concept of density peaks to select the examples to save in the replay buffer $\mathcal{R}$.  
In~\cref{tab:centroid_selection} we compare centroid selection and our proposed density-based selection. 
We observe that our method achieves superior performance.

\begin{table}[h]
\centering
\resizebox{0.85\columnwidth}{!}{
\begin{tabular}{l |rr | rr}
\toprule
Methods        & \multicolumn{2}{c}{CIFAR-100} & \multicolumn{2}{|c}{iNatIGCD-l} \\
  & \multicolumn{1}{c}{$M_f \downarrow$} & \multicolumn{1}{c}{$M_d \uparrow$} & \multicolumn{1}{|c}{$M_f \downarrow$} & \multicolumn{1}{c}{$M_d \uparrow$} \\
\midrule
Ours w/ iCaRL~\cite{rebuffi2017icarl} &  \textbf{3.4}          & \uline{30.7}          &  \uline{28.3}      & \uline{27.1}       \\
Ours                                   & \uline{4.0}   & \textbf{31.2} & \textbf{27.6}        & \textbf{28.4}      \\
\bottomrule
\end{tabular}
}
\vspace{-5pt}
\caption{
Impact of the replay buffer selection method. 
}
\label{tab:centroid_selection}
\end{table}

\noindent{\bf Impact of the Size of $\mathcal{S}$.}
One of the key factors influencing the performance of our $\texttt{SNN}$ classifier is the size of $\mathcal{S}$. 
In~\cref{tab:size_of_s}, we study the influence of this hyper-parameter by varying the number of examples per category ($N^\mathcal{S}$/$K^{\mathcal{S}}$). 
When only using one example per category, our model exhibits  forgetting and inferior discovery performance. 
Increasing the number of examples per category increases the performance, until it starts to plateau after five examples per category. 
To reduce the cost of saving too many examples to $\mathcal{S}$, we  set $N^\mathcal{S}$/$K^\mathcal{S}$ to five in our experiments.

\begin{table}[h]
\centering
\resizebox{0.7\columnwidth}{!}{
\begin{tabular}{c |rr | rr}
\toprule
 $N^{\mathcal{S}}$ / $K^{\mathcal{S}}$ & \multicolumn{2}{c}{CIFAR-100} & \multicolumn{2}{|c}{iNatIGCD-l} \\
  & \multicolumn{1}{c}{$M_f \downarrow$} & \multicolumn{1}{c}{$M_d \uparrow$} & \multicolumn{1}{|c}{$M_f \downarrow$} & \multicolumn{1}{c}{$M_d \uparrow$} \\
\midrule
1               &    20.4     &    16.7    & 39.6  & 15.7 \\
3               &    8.5      &    25.7    & 31.2  & 20.5 \\
5               &    4.0      &    31.2    & 27.6  & 28.4  \\
7               &    \uline{4.1}       &      \uline{32.0}          &   \uline{27.0}        &   \uline{28.6}       \\
10              &    \textbf{3.1}      &    \textbf{33.2}    & \textbf{25.1}  & \textbf{29.5} \\
\bottomrule
\end{tabular}
}
\vspace{-5pt}
\caption{
Impact of the size of the support set $\mathcal{S}$. 
}
\label{tab:size_of_s}
\end{table}

\section{Conclusion}
We explored the problem of incremental generalized category discovery (IGCD). 
To do this, we constructed a new dataset, iNatIGCD, motivated by a real-world fine-grained visual discovery task and used it to benchmark the performance of recent category discovery methods. 
Through our experiments, we showed that our new approach which combines non-parametric categorization with a density-based sample selection technique is superior to existing methods. 
While promising, our results show that the IGCD problem, especially in our large-scale fine-grained setting, remains challenging for existing methods. 
We hope that our work opens the door to future progress on this task.

\noindent{\bf Acknowledgements.} We thank the iNaturalist community for their data collection efforts. 

{\small
\bibliographystyle{ieee_fullname}
\bibliography{main}
}

\clearpage
\appendix
\setcounter{table}{0}
\renewcommand{\thetable}{A\arabic{table}}
\setcounter{figure}{0}
\renewcommand{\thefigure}{A\arabic{figure}}

\begin{figure*}
\begin{center}
\centering
\captionsetup{type=figure}
    \includegraphics[width=1.0\textwidth]{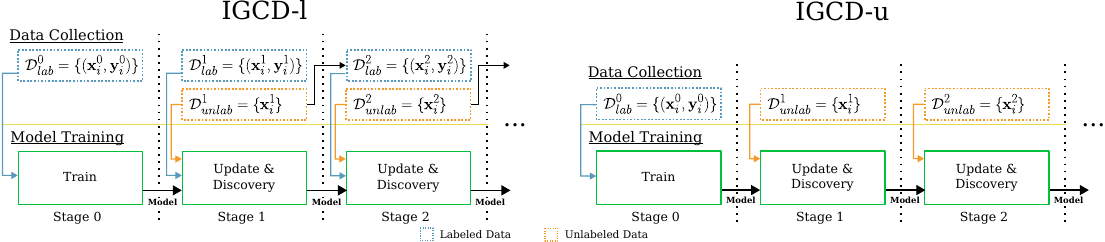}
\caption{
The two incremental learning settings used in our work. In the \mbox{IGCD-l} setting, at each stage we have one unlabeled set and one labeled set.  
The unlabeled set is annotated at the end of each stage and then used as the labeled set for the next stage.
In the IGCD-u setting, only the unlabeled set is provided at each incremental stage, \ie after the initial label data is given in Stage-0, no additional labels are provided. 
}
\label{fig:igcd_setting}
\end{center}
\end{figure*}

\section{iNatIGCD Dataset}

In~\cref{tab:dataset_stats}, we compare our dataset split to common benchmarks used by previous papers. 
Our new iNatIGCD benchmark has more categories and images per stage and thus can be used to better evaluate the performance of generalized category discovery methods.

\section{Additional Results}

\begin{table*}[t]
    \centering
    \resizebox{\textwidth}{!}{
    \begin{tabular}{l l c c c c}
    \toprule
          Dataset      & Used In                                    & \# Classes / Stage             & \# Stages & \# Avg. Images / Stage & \# Avg. Classes / Stage\\ \midrule
          ImageNet-1k & NCDwF~\cite{joseph2022novel}       &  882 / 30                &   2       &   39k             &   30              \\
          TinyImageNet  & class-iNCD~\cite{roy2022class}  &  180 / 20                &   2       &   20k             &   20              \\
          CIFAR100-MI & GM~\cite{zhang2022grow}            &  70 / 10 / 10 / 10           &   4       &   5k              &   10              \\ \midrule
          iNatIGCD & Ours         &  3,888 / 972 / 3,040 / 4,324    &   4       &   25k             &   2.7k \\ \bottomrule
    \end{tabular}
    }
   \vspace{-5pt}
    \caption{
    The statistics of our dataset compared to previous benchmarks that focus on generalized category discovery under the incremental setting. At Stage-1, 2, and 3, iNatIGCD contains 390, 2,625, and 1,492 novel classes respectively.
    }
   \vspace{-5pt}
    \label{tab:dataset_stats}
\end{table*}

\subsection{Additional Ablations}

\noindent{\bf Impact of the Size of $\mathcal{R}$.}
In addition to the ablation on the size of the support set $\mathcal{S}$ in the main paper, here we  present results where we vary the size of the replay buffer $\mathcal{R}$.
Similar to the ablation of the size of $\mathcal{S}$, we denote the number of images in $\mathcal{R}$ as $N^\mathcal{R}$ and the number of categories in $\mathcal{R}$ as $K^{\mathcal{R}}$.
The results are presented in~\cref{tab:size_of_r}. 
The default value used by our method is illustrated by the gray row, \ie the second row. 
We observe a similar trend as Table~6 in the main paper where increasing the number of examples increases the performance.  
However, the larger number impacts training efficiency especially in the context of the large number of categories in our iNatIGCD dataset.

\begin{table}[h]
\centering
\resizebox{0.7\columnwidth}{!}{
\begin{tabular}{c |rr | rr}
\toprule
 $N^{\mathcal{R}}$ / $K^{\mathcal{R}}$ & \multicolumn{2}{c}{CIFAR-100} & \multicolumn{2}{|c}{iNatIGCD-l} \\
  & \multicolumn{1}{c}{$M_f \downarrow$} & \multicolumn{1}{c}{$M_d \uparrow$} & \multicolumn{1}{|c}{$M_f \downarrow$} & \multicolumn{1}{c}{$M_d \uparrow$} \\
\midrule
1               &    25.1              &    18.6                     & 38.2  & 15.8 \\
\baseline{3}    &    \baseline{4.6}    &    \baseline{29.2}    & \baseline{28.3}  & \baseline{27.1} \\
5               &    \uline{4.0}      &    31.2             & 28.1  & 26.6  \\
7               &    4.4       &      \uline{34.2}          &   \uline{27.1}   &   \uline{27.9}       \\
10              &    \textbf{3.4}      &    \textbf{36.1}    & \textbf{24.0}  & \textbf{29.4} \\
\bottomrule
\end{tabular}
}
\vspace{-5pt}
\caption{
Impact of the size of the replay buffer $\mathcal{R}$. 
}
\vspace{-10pt}
\label{tab:size_of_r}
\end{table}

\noindent{\bf CLIP pretrained features.}
We also performed ablation experiments on iNatIGCD-l using CLIP pre-trained feature extractor. 
In~\cref{tab:clip_res}, we can see that due to the challenging fine-grained nature of iNatIGCD, the model failed to perform discovery well. 
But overall the pre-trained CLIP feature extractor improves the performance, and our method still achieves the best performance.

\begin{table}[h]
\centering
\resizebox{0.90\columnwidth}{!}{
\begin{tabular}{lccc}
\toprule
$M_f\downarrow / M_d\uparrow$ & CLIP-RN-50 & RN-18       & ViT-B     \\
\midrule
GM~\cite{zhang2022grow}        & \uline{28.4} / \uline{25.6} & \uline{30.0} / \uline{20.1} & \textbf{29.1} / \uline{24.8} \\
Ours      & \textbf{25.4} / \textbf{34.2} & \textbf{27.6} / \textbf{28.4} & \uline{29.4} / \textbf{30.2} \\
\bottomrule
\end{tabular}
}
\vspace{-5pt}
\caption{Performance of using CLIP pretrained features on iNAT-IGCD-l dataset.}
\label{tab:clip_res}
\end{table}

\noindent{\bf Density Selection Parameters.}
There are three hyper-parameters to be set in our density selection step described in Sect.~4.3 in the main paper.  
They are, the number of nearest neighbors $K$ for calculating the density, the number of neighbors $K^d$, and the threshold $T$ used when removing redundant density peaks.
We present the ablation study on these parameters in~\cref{tab:density_k,tab:density_k_d,tab:density_t}.

In~\cref{tab:density_k}, we vary the value of $K$. 
We can see that generally the performance of different $K$ values results in an inverse U shape. 
As $K$ increases, the performance reaches the best value, and after the performance peak, the performance degrades as $K$ increases further. 
When $K$ is small, the compared neighborhood is small, thus the estimation of density peaks is not accurate. 
This results in many noisy peaks that can lead worse performance. 
When $K$ increases, the density peak estimation is more accurate. 
However, considering the case where $K$ equals to the number of images in the dataset, we would underestimate the density peaks as many clusters may be considered as one cluster. 
This would result in a performance decrease when $K$ is higher than better choices.

\begin{table}[h]
\centering
\resizebox{0.7\columnwidth}{!}{
\begin{tabular}{c |rr | rr}
\toprule
 $K$ & \multicolumn{2}{c}{CIFAR-100} & \multicolumn{2}{|c}{iNatIGCD-l} \\
  & \multicolumn{1}{c}{$M_f \downarrow$} & \multicolumn{1}{c}{$M_d \uparrow$} & \multicolumn{1}{|c}{$M_f \downarrow$} & \multicolumn{1}{c}{$M_d \uparrow$} \\
\midrule
5               &    4.8               &    27.1                       & 28.9             & 27.1 \\
\baseline{10}   &    \baseline{4.3}    &    \baseline{\uline{28.9}}    & \baseline{27.9}  & \baseline{26.9} \\
15              &    4.5               &    \textbf{29.7}              & \uline{26.1}     & \textbf{27.8}  \\
20              &    \uline{4.1}       &    28.1                       & \textbf{25.9}    & \uline{27.6}       \\
40              &    \textbf{3.8}      &    26.4                       & 27.2             & 26.0 \\
\bottomrule
\end{tabular}
}
\vspace{-5pt}
\caption{
Ablation of $K$ used in density-based selection. 
}
\vspace{-10pt}
\label{tab:density_k}
\end{table}

In~\cref{tab:density_k_d}, we ablate the choice of $K^d$. 
This hyper-parameter is used to remove redundant density peaks. 
When the value of $K^d$ is low, it would remove fewer density peaks as two peaks' $K^d$ neighbor set are less likely to overlap.
When the value is high, it would remove more density peaks. 
From the results, we can see that setting $K^d$ to a higher number than $K$ (which equals $10$ in~\cref{tab:density_k_d}) can result in better performance.

\begin{table}[h]
\centering
\resizebox{0.7\columnwidth}{!}{
\begin{tabular}{c |rr | rr}
\toprule
 $K^d$ & \multicolumn{2}{c}{CIFAR-100} & \multicolumn{2}{|c}{iNatIGCD-l} \\
  & \multicolumn{1}{c}{$M_f \downarrow$} & \multicolumn{1}{c}{$M_d \uparrow$} & \multicolumn{1}{|c}{$M_f \downarrow$} & \multicolumn{1}{c}{$M_d \uparrow$} \\
\midrule
10               &    \uline{4.1}                 &    20.3                    & 28.0             & 18.9 \\
15               &    4.8                 &    23.2                    & 27.5             & 23.4 \\
\baseline{20}    &    \baseline{4.3}      &    \baseline{\uline{28.9}}         & \baseline{27.9}  & \baseline{26.9}  \\
30               &    \textbf{3.8}                 &    \textbf{29.5}                    & \uline{27.1}             & \textbf{27.9}       \\
40               &    4.5                 &    28.1                    & \textbf{26.9}             & \uline{27.2} \\
\bottomrule
\end{tabular}
}
\vspace{-5pt}
\caption{
Ablation of $K^d$ used in density-based selection. 
}
\vspace{-10pt}
\label{tab:density_k_d}
\end{table}

Another hyper-parameter in density selection is the threshold $T$. 
Similar to the selection of $K^d$, a higher threshold will remove fewer density peaks, while a lower threshold will remove more. 
The results are presented in~\cref{tab:density_t}. 
Our choice from the main paper is shaded in gray, which obtains a good balance between forgetting and discovery.

\begin{table}[h]
\centering
\resizebox{0.7\columnwidth}{!}{
\begin{tabular}{c |rr | rr}
\toprule
 $T$ & \multicolumn{2}{c}{CIFAR-100} & \multicolumn{2}{|c}{iNatIGCD-l} \\
  & \multicolumn{1}{c}{$M_f \downarrow$} & \multicolumn{1}{c}{$M_d \uparrow$} & \multicolumn{1}{|c}{$M_f \downarrow$} & \multicolumn{1}{c}{$M_d \uparrow$} \\
\midrule
0.2               &  5.1             &    28.7             & 29.2             & \textbf{27.0} \\
0.4               &  4.8             &    \textbf{29.3}             & 28.9             & 26.8 \\
\baseline{0.6}    & \baseline{\uline{4.3}}   &    \baseline{\uline{28.9}}  & \baseline{\uline{27.9}}  & \baseline{\uline{26.9}}  \\
0.8               &  \textbf{4.0}             &    27.2             & \textbf{27.1}             & 26.0       \\
\bottomrule
\end{tabular}
}
\vspace{-5pt}
\caption{
Ablation of $T$ used in density-based selection.
}
\vspace{-10pt}
\label{tab:density_t}
\end{table}

\subsection{Vision Transformer Results}

In the main paper, our results are presented using a ResNet-18 backbone. 
In~\cref{tab:vit_inat_IGCDL,tab:vit_inat_IGCDU} we present results using a vision transformer~\cite{dosovitskiy2020vit} Base 16 model. 
We observe that the overall trend remains the same. 
The vision transformer results in an increase in performance compared to the ResNet-18~\cite{he2016deep} that we use in the main paper, but at the cost of being much slower to train.

\begin{table*}[t]
\centering
\resizebox{0.95\textwidth}{!}{
\begin{tabular}{l | c | ccc | cccc | ccccc | cc}
\toprule
Methods                                 & Stage-0      & \multicolumn{3}{|c}{Stage-1}  & \multicolumn{4}{|c}{Stage-2}   & \multicolumn{5}{|c}{Stage-3}    &  \multicolumn{2}{|c}{Overall}    \\ 
\midrule
                                         & All          & All  & Old  & New           & All  & Old  & New    & S-0    & All  & Old  & New   & S-1  & S-0   & $M_f$     &  $M_d$  \\
\midrule
SimGCD + iCaRL~\cite{rebuffi2017icarl}   & 45.6  & \uline{25.2} & \uline{33.6} & \uline{14.1}      & \uline{24.7} & \textbf{38.0} & \uline{16.9}  & 9.7    & \uline{24.0} & \uline{37.0} & \uline{17.8}  & 15.3 & 9.7  & 35.9 & \uline{26.9}    \\
GM~\cite{zhang2022grow}                  & \textbf{46.0}  & 19.5 & 28.1 & 9.4       & 20.9 & 29.6 & 11.2   & \uline{16.2}   & 18.3 & 27.1 & 16.8  & \uline{17.0} & \uline{14.3} & \textbf{29.1} & 24.8 \\
\midrule
Ours                                     & 45.6  & \textbf{27.0} & \textbf{36.3} & \textbf{16.7}      & \textbf{26.8} & \uline{37.2} & \textbf{23.4}   & \textbf{18.5}   & \textbf{25.0} & \textbf{38.1} & \textbf{19.7}  & \textbf{18.2} & \textbf{17.1} & \uline{29.4} & \textbf{30.2}    \\
\bottomrule
\end{tabular}
}
\vspace{-5pt}
\caption{
Vision transformer-backbone results on iNatIGCD in the IGCD-l setting (\ie where labels are available at the end of each stage). Higher numbers are better, with the exception of $M_f$, where lower is better. 
}
\label{tab:vit_inat_IGCDL}
\end{table*}

\begin{table*}[t]
\centering
\resizebox{0.8\textwidth}{!}{
\begin{tabular}{l | c | cc | ccc | cccc | cc}
\toprule
Methods                                 & Stage-0      & \multicolumn{2}{|c}{Stage-1}  & \multicolumn{3}{|c}{Stage-2}   & \multicolumn{4}{|c}{Stage-3}    &  \multicolumn{2}{|c}{Overall}    \\ 
\midrule
                                         & All          & New & S-1 & New           & S-1 & S-0 & New    & S-2    & S-1 & S-0    & $M_f$     &  $M_d$  \\
\midrule
SimGCD + iCaRL~\cite{rebuffi2017icarl} & 42.0     &  \uline{9.8}     &  17.9     &  \uline{14.6} & 8.9  &  12.6  &  \uline{15.7}     &   11.2     & 7.4      &  10.4      &  31.3   & \uline{12.6} \\
GM~\cite{zhang2022grow}    & \textbf{42.1}     & 9.0      &  \uline{18.0}    &  14.0      &  \uline{9.4}    &  \textbf{15.4}     &   10.8    &   \uline{11.3}     &   \uline{8.4}    &    \uline{13.1}    &  \uline{29.2}   & 10.6 \\
\midrule
Ours                & 41.9         & \textbf{12.8} & \textbf{22.5}           & \textbf{15.7} & \textbf{10.9} & \uline{14.6}      & \textbf{16.3} & \textbf{11.5} & \textbf{9.0}  & \textbf{13.6} & \textbf{28.4} & \textbf{14.9} \\
\bottomrule
\end{tabular}
}
\vspace{-5pt}
\caption{
Vision transformer-backbone results on iNatIGCD in the IGCD-u setting (\ie where labels are not available at the end of each stage). Higher numbers are better, with the exception of $M_f$, where lower is better. 
}
\label{tab:vit_inat_IGCDU}
\end{table*}

\subsection{CIFAR-100 IGCD-l and IGCD-u Results}
In~\cref{tab:cifar_100_igcdl,tab:cifar_100_igcdu}, we re-purpose the CIFAR-100 dataset for the IGCD-l and IGCD-u settings, using the same category split as~\cite{zhang2022grow}, and present the results comparing our method with previous SoTA baselines. 
We can see that our method, when compare with others, still achieves a competitive performance. 
In terms of the overall performance metrics $\mathcal{M}_f$ and $\mathcal{M}_d$, our method achieves the best results for both CIFAR-100-IGCD-l and CIFAR-100-IGCD-u.

\begin{table*}[t]
\centering
\resizebox{0.95\textwidth}{!}{
\begin{tabular}{l | c | ccc | cccc | ccccc | cc}
\toprule
Methods                                 & Stage-0      & \multicolumn{3}{|c}{Stage-1}  & \multicolumn{4}{|c}{Stage-2}   & \multicolumn{5}{|c}{Stage-3}    &  \multicolumn{2}{|c}{Overall}    \\ 
\midrule
                                         & All          & All  & Old  & New           & All  & Old  & New    & S-0    & All  & Old  & New   & S-1  & S-0   & $M_f$     &  $M_d$  \\
\midrule
Supervised upper-bound                   & 79.2         & 82.5 & 80.3 & 85.3           & 81.0 & 82.4 & 80.0   & 79.2   & 81.3 & 82.3 & 80.2  & 81.7 & 80.2  & - & - \\ 
\midrule
SimGCD + iCaRL~\cite{rebuffi2017icarl}   & 77.8         & \textbf{78.6} & \textbf{79.6} & 76.7           & 77.9 & 78.6 & 74.5   & 63.4   & 75.6 & \textbf{78.9} & 70.8  & 60.2 & \uline{50.1}  & \uline{27.7} & 67.8   \\
GM~\cite{zhang2022grow}                  & 79.0         & 78.3 & 78.8 & \uline{77.3}           & \textbf{78.2} & \textbf{78.9} & \uline{74.7}   & \textbf{66.3}   & \textbf{76.7} & 77.8 & \uline{71.2}  & \uline{63.2} & 49.8 & 29.2 & \uline{68.3} \\
\midrule
Ours                                     & 77.8         & \uline{78.5} & \uline{79.0} & \uline{78.2}           & \uline{78.0} & \uline{78.7} & \textbf{75.0}   & \uline{65.3}   & \uline{76.0} & \uline{78.8} & \textbf{72.1}  & \textbf{64.6} & \textbf{51.3} & \textbf{26.5} &  \textbf{69.2}   \\
\bottomrule
\end{tabular}
}
\vspace{-5pt}
\caption{
Results on CIFAR-100 in the IGCD-l setting (\ie where labels are available at the end of each stage). Higher numbers are better, with the exception of $M_f$, where lower is better.
}
\vspace{-5pt}
\label{tab:cifar_100_igcdl}
\end{table*}

\begin{table*}[h]
\centering
\resizebox{0.82\textwidth}{!}{
\begin{tabular}{l | c | cc | ccc | cccc | cc}
\toprule
Methods                                & Stage-0      & \multicolumn{2}{|c}{Stage-1}  & \multicolumn{3}{|c}{Stage-2}   & \multicolumn{4}{|c}{Stage-3}    & \multicolumn{2}{|c}{Overall}    \\ 
\midrule
                                       & All          & New & S-0          & New & S-1 & S-0        &  New &  S-2 & S-1 & S-0  & $M_f$   &  $M_d$  \\
\midrule
Supervised upper-bound                 & 79.2         & 78.9 & 81.2        & 80.2 & 79.9 & 81.4     & 85.3  & 82.1 & 80.1  & 79.0  & -  & - \\ 
\midrule
SimGCD + iCaRL~\cite{rebuffi2017icarl} & 77.8         & \uline{62.4} & 43.2        & \uline{66.7} & 40.8 & 38.7     & \textbf{68.2}  & \uline{41.3} & 36.2  &  35.3 & 42.5  & 50.1 \\
GM~\cite{zhang2022grow}                & 79.0         & 58.9 & \uline{45.3}        & 63.4 & \uline{46.8} & \textbf{43.2}     & 60.0  & \textbf{42.5} & \uline{45.6}  & \uline{40.2} & \uline{38.8}  & \uline{53.4} \\
\midrule
Ours                                   & 77.8         & \textbf{64.4} & \textbf{46.9}        & \textbf{67.8} & \textbf{47.0} & \uline{42.1}     & \uline{68.0}  & 40.0 & \textbf{46.8}  & \textbf{41.5} & \textbf{36.3} & \textbf{55.9} \\
\bottomrule
\end{tabular}
}
\vspace{-5pt}
\caption{
Results on CIFAR-100 in the IGCD-u setting (\ie where labeled data is not provided during the incremental stages). 
Higher numbers are better, with the exception of $M_f$, where lower is better. 
}
\vspace{-5pt}
\label{tab:cifar_100_igcdu}
\end{table*}

\subsection{Results on Static GCD Datasets}

In \cref{tab:static_gcd} we present results on the static GCD benchmarks, \ie without any incremental stages. 
For the adaptation of our method to the static GCD scenario, we remove the incremental update of $\mathcal{S}$ and $\mathcal{R}$. 
We can see that our proposed method still performs on par with previous SoTA methods on static GCD benchmarks despite primarily being designed for the incremental setting.

\begin{table}[h]
\resizebox{1.0\columnwidth}{!}{
\begin{tabular}{l|lll|lll}
\toprule
Methods       & \multicolumn{3}{|c}{ImageNet-100}       & \multicolumn{3}{|c}{SCars} \\ 
                         & All     & Old  & New  & All     & Old    & New    \\ 
\midrule
GCD~\cite{vaze2022generalized} & 74.1    & 89.8  & 66.3 & 39.0   &  57.6 & 29.9   \\
ORCA~\cite{cao22}              & 73.5    & \textbf{92.6}  & 63.9 & 23.5   & 50.1 & 10.7 \\
SimGCD~\cite{wen2023simple}    & \uline{82.4}  & \uline{90.7} & \uline{78.3} & \uline{46.8}    & \textbf{64.9}   & \uline{38.0}   \\
\midrule
Ours                           & \textbf{83.0} & 89.5 & \textbf{79.1} & \textbf{47.2}    & \uline{63.8}   & \textbf{38.7}   \\ 
\bottomrule
\end{tabular}
}
\vspace{-5pt}
\caption{Results on static GCD benchmarks. 
}
\vspace{-10pt}
\label{tab:static_gcd}
\end{table}

\section{Implementation Details}

\subsection{Training Losses}
We describe the representation learning losses $\mathcal{L}_{\text{rep}}$ below which follows a contrastive learning framework. 
Formally, given two views (\ie augmentations) $\hat{\bm{x}}_i$ and $\tilde{\bm{x}}_i$ of an input image $\bm{x}_i$ in a mini-batch $\mathcal{B}$, the self-supervised component of the contrastive loss can be written as
\begin{equation}
    \mathcal{L}_{\text{SelfCon}} = \frac{1}{|
\mathcal{B}|} \sum_{i \in \mathcal{B}}-\log \frac{\exp \left(\hat{\bm{z}}_i^\top \tilde{\bm{z}}_i^{\prime} / \tau_u\right)}{\sum_i^{i \neq n} \exp \left(\hat{\bm{z}}_i^\top \tilde{\bm{z}}_n^{\prime} / \tau_u\right)} \,,
\end{equation}
where the embedding $\bm{z}=m(f(\bm{x}))$ is a projected feature from a MLP projector $m$ as in~\cite{chen2020simple,wen2023simple}, and $\tau_u$ is a temperature parameter.
The supervised contrastive loss~\cite{khosla2020supervised} is similar with the difference being that the positive samples are matched with their ground truth labels,
\begin{equation}
    \mathcal{L}_{\text{SupCon}} = \frac{1}{|\mathcal{B}^l|} \sum_{i \in \mathcal{B}^l}\frac{1}{|\mathcal{M}_i|} \sum_{q \in \mathcal{M}_i} -\log \frac{\exp \left(\hat{\bm{z}}_i^\top \tilde{\bm{z}}_q / \tau_c\right)}{\sum_n^{n \neq i} \exp \left(\hat{\bm{z}}_i^\top \tilde{\bm{z}}_n / \tau_c\right)} \,,
\end{equation}
where $\mathcal{M}_i$ indexes all other images in the same batch that have the same label as $\bm{x}_i$, and $\tau_c$ is the temperature parameter for the supervised contrastive loss.

For the fully supervised upper bound discussed in Tabs.~1 and 2 in the main paper, we leverage both contrastive learning losses $\mathcal{L}_{\text{SelfCon}}$ and $\mathcal{L}_{\text{SupCon}}$ for representation learning. 
For the classifier, we simply adopt the cross-entropy loss $\mathcal{L}_{\text{ce}}$ on all examples.
We concatenate all datasets $\mathcal{D}_{lab}^t, \mathcal{D}_{unlab}^t$ from every time step~$t$, and provide the ground truth category labels for the unlabelled datasets $\mathcal{D}_{unlab}^t$. 
The resulting concatenated dataset is used to train the fully supervised baseline.
So in this setting, the model will have labels for all the categories and no forgetting occurs. 
This can be considered as a very strong upper bound on the performance for our IGCD setting.

\begin{figure}
    \centering
    \includegraphics[width=\linewidth]{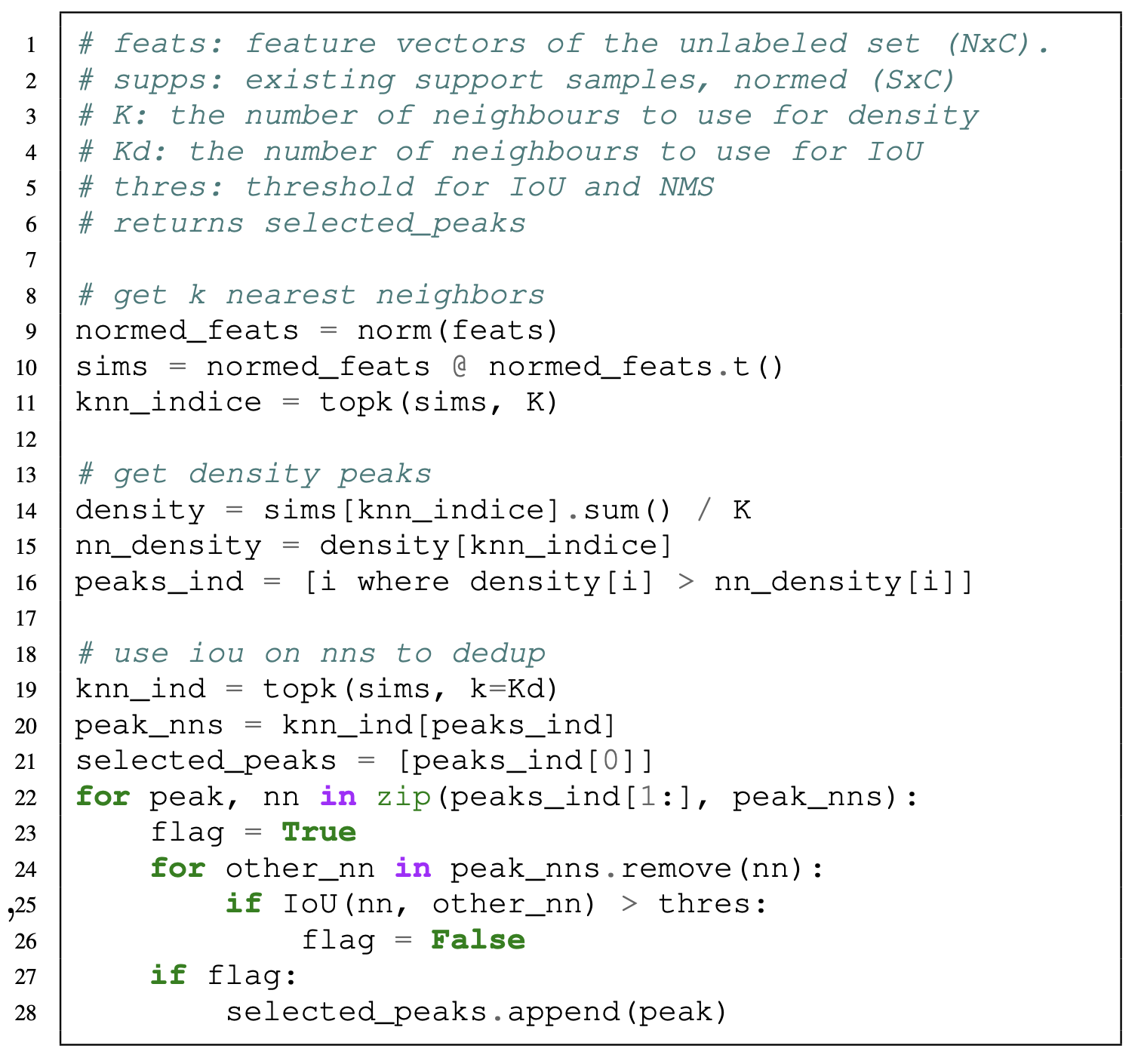}
\caption{
Pseudo-code of our density selection step.  
}
\vspace{-5pt}
\label{alg:density_select_code}
\end{figure}

\subsection{Density Selection}

We present the pseudo-code for our proposed density selection process in~\cref{alg:density_select_code}. 
We use $\ell_2$-normed features to calculate the density of the examples, and then filter out the density peaks by comparing the density of one node to its $k$ nearest neighbours.
Then we compare each pair of density peaks and remove peaks that have nearest neighbours set to overlap with other peaks larger than a threshold.

\subsection{Evaluation Metrics}
The main results in each of the tables and figures are reported using the clustering accuracy (ACC).
At each evaluation stage, given the ground truth $\bm{y}$ and a predicted label $\hat{\bm{y}}$, the ACC is calculated as $\text{ACC} = \frac{1}{M} \sum_{i=1}^{M} \mathds{1}(\bm{y}_i = p(\hat{\bm{y}}_i))$. 
Here, $M = |\mathcal{D}^u|$ and $p$ is the optimal permutation that matches the predicted cluster assignments to the ground truth category labels.
Under the IGCD-l setting, we report the ACC for all the categories in Stage-$0$ as all categories have labeled training examples. 
For later stages, we report the `All', `Old', and `New' performance which correspond to all categories, the labeled categories, and the unlabeled categories at each stage.
Additionally, we use S-$t$ to denote the categories that appear in a previous stage $t$ but are not presented in the current stage to measure the forgetting of the model.
Finally, as noted in the main paper, we also use the summary metrics defined in~\cite{zhang2022grow} to measure overall discovery and forgetting performance, $\mathcal{M}_d$ and $\mathcal{M}_f$.

\subsection{Experimental Settings}

Our models are trained using a stochastic gradient descent optimizer with an initial learning rate  of $0.1$,  momentum of $0.9$, and weight decay of $1e-4$. 
The learning rate is decayed following a cosine schedule~\cite{loshchilov2016sgdr}. 
We use a batch size of $128$ during training for all datasets, where $64$ images are sampled from $\mathcal{D}_{lab}^t$ and $64$ images are  sampled from $\mathcal{D}_{unlab}^t$.
The balancing factor $\lambda_{\text{rep}}$ is set to $0.35$, $\epsilon$ is set to $2.0$, $K$ is set to $10$, $K^d$ is set to $20$, and the threshold $T$ is set to $0.6$. 
We adopt the same set of augmentations used in~\cite{uno} for contrastive learning. 
The temperature parameters in contrastive learning $\tau_u$ and $\tau_c$ are set to $0.07$ and $0.1$ respectively.

\end{document}